\documentclass[10pt,journal]{IEEEtran}

\usepackage{cite}
\usepackage{amsmath,amssymb,amsfonts}
\usepackage{algorithmic}
\usepackage{graphicx}
\usepackage{textcomp}
\usepackage{hyperref}
\usepackage{booktabs}
\usepackage{multirow}
\usepackage{url}
\usepackage{xcolor}
\usepackage{array}
\usepackage{subcaption}

\def\BibTeX{{\rm B\kern-.05em{\sc i\kern-.025em b}\kern-.08em
    T\kern-.1667em\lower.7ex\hbox{E}\kern-.125emX}}

\newcommand{\paragraphX}[1]{\vskip 4pt \noindent \textit{#1} \hskip .05in}

\usepackage[color=green]{todonotes}

\begin{document}

\title{Towards Quantum Machine Learning for Malicious Code Analysis}


\author{
\IEEEauthorblockN{
Jesus Lopez\IEEEauthorrefmark{2}, 
Saeefa Rubaiyet Nowmi\IEEEauthorrefmark{2}, 
Viviana Cadena\IEEEauthorrefmark{2}, 
and Mohammad Saidur Rahman\IEEEauthorrefmark{2}
}\\
\IEEEauthorblockA{
\IEEEauthorrefmark{2}Department of Computer Science, University of Texas at El Paso, TX, USA\\
Email: \texttt{\{jlopez126,snowmi,vcadena1\}@miners.utep.edu}, \texttt{msrahman3@utep.edu}
}
}

\maketitle

\begin{abstract}

Classical machine learning (CML) has been extensively studied for malware classification. With the emergence of quantum computing, quantum machine learning (QML) presents a paradigm-shifting opportunity to improve malware detection, though its application in this domain remains largely unexplored. In this study, we investigate two hybrid quantum-classical models --- a Quantum Multilayer Perceptron (QMLP) and a Quantum Convolutional Neural Network (QCNN), for malware classification. Both models utilize \emph{angle embedding} to encode malware features into quantum states. QMLP captures complex patterns through full qubit measurement and data re-uploading, while QCNN achieves faster training via quantum convolution and pooling layers that reduce active qubits. We evaluate both models on five widely used malware datasets --- API-Graph, EMBER-Domain, EMBER-Class, AZ-Domain, and AZ-Class, across binary and multiclass classification tasks.

Our results show high accuracy for binary classification --- 95–96\% on API-Graph, 91–92\% on AZ-Domain, and 77\% on EMBER-Domain. In multiclass settings, accuracy ranges from 91.6–95.7\% on API-Graph, 41.7–93.6\% on AZ-Class, and 60.7–88.1\% on EMBER-Class. Overall, QMLP outperforms QCNN in complex multiclass tasks, while QCNN offers improved training efficiency at the cost of reduced accuracy.

\end{abstract}


\textbf{Availability.} The implementation is available at \url{https://iqsec-lab.github.io/qml-malware_classification/}.

\section{Introduction}


Quantum Machine Learning (QML) is an emerging, interdisciplinary field of research that seeks to solve complex problems across a wide range of domains by leveraging the power of quantum computing \cite{ nguyen2025diffusion, cerezo2022challenges, biamonte2017quantum}. Within this field, one of the most actively studied areas is Quantum Neural Networks (QNNs)~\cite{biamonte2017quantum}. QNNs are quantum analogs of classical neural networks; instead of neurons and weighted connections, they use quantum circuits composed of parameterized gates to learn from quantum-encoded inputs~\cite{schuld2014quest,abbas2021power}.

While QNNs are promising, many QNNs are limited by Noisy Intermediate-Scale Quantum (NISQ) devices\cite{Preskill2018}, which suffer from high error rates, limited qubit counts and shallow circuit depths \cite{hur2021quantum}. Despite these limitations, QNNs have shown potential in learning complex quantum-encoded patterns with fewer parameters than classical networks. As a result, current research efforts focus on building QNN architectures that are more robust and better suited to NISQ-era devices.

Most QNN models have primarily focused on MNIST image classification~\cite{wang2022, wang2021quantumnat}. MNIST provides benchmarks for testing models that are relatively easier to encode into quantum circuits. However, due to the current hardware limitations, the full $28\times28$ images cannot be encoded directly and have to be downsized to more feasible dimensions, such as $4\times4$ in order to fit the number of qubits used in a quantum circuit~\cite{qnn2022evaluation}. Quantum simulators such as PennyLane~\cite{PennyLane} and TensorFlow Quantum~\cite{tensorflow-quantum} struggle with large datasets as they have to compute each sample in a circuit \cite{Preskill2018}. Despite the significant reduction in resolution, the model still yields promising results even when MNIST images are downsized to $4\times4$, demonstrating the robustness of the approach~\cite{qmlp,wang2022}.

Even though QML has been widely studied in domains such as computer vision \cite{qmlp}, chemistry \cite{sajjan2022quantum}, and material discovery \cite{choudhary2019accelerated}, its applicability and effectiveness in malicious code analysis remain largely unexplored. As such, in this work, we explore two of the most widely adapted QML models --- Quantum Convolutional Neural Networks (QCNN) \cite{qcnn} and Quantum Multilayer Perceptrons (QMLP) \cite{qmlp}, to assess their broader applicability and effectiveness for malicious code (i.e., malware) classification. Unlike images, malware datasets often contain high-dimensional binary feature vectors derived from static characteristics such as API calls and their relationships, permissions, and intent filtering \cite{ember, api_graph_dataset, arp2014drebin}. 
We have implemented both Quantum Multilayer Perceptron (QMLP) and Quantum Convolutional Neural Network (QCNN) architectures within the PennyLane \cite{PennyLane} simulator. These models, originally developed for image classification, are evaluated on five benchmark malware datasets --- API-Graph \cite{api_graph_dataset}, EMBER-Class \cite{continual-learning-malware}, EMBER-Domain \cite{ember}, AZ-Class \cite{malcl, madar}, and AZ-Domain \cite{malcl, madar}. These datasets comprise both Windows \cite{ember, continual-learning-malware} and Android \cite{api_graph_dataset, malcl, madar} malicious and benign applications. Crucially, their feature sets contain various API-call graphs, offering a more complex and structured input format compared to traditional image data.

In summary, the contributions of this paper are as follows:
\begin{itemize}
    \item We extensively evaluated Quantum Multilayer Perceptron (QMLP) and Quantum Convolutional Neural Network (QCNN) models across five diverse malware datasets, encompassing both multi-class and binary classification scenarios.
    \item This study provides an in-depth analysis of how the architectural design choices of QMLP and QCNN impact their accuracy and overall performance in malware classification across various datasets.
    \item We observe that the QCNN model, by leveraging convolution and pooling mechanisms, offers improved time efficiency. However, this comes at the cost of reduced accuracy across most datasets except for the API-Graph dataset, where QCNN slightly outperforms QMLP in multiclass classification.
    
\end{itemize}

\section{Preliminaries}
In this section, we discuss the details of architectures and the workflows of the Quantum Multilayer Perceptron (QMLP) and Quantum Convolutional Neural Network (QCNN) used in our experiments.

\subsection{QMLP} 

The major two components of QMLP are --- a parameterized quantum circuit (called a QNode) and a classical fully connected layer that produces the final output. The quantum circuit consists of 16 qubits representing one of the principal component from the malware dataset. The input features are encoded into the quantum system using $RX$ rotation gates, which convert classical values into quantum states. After the data encoding, the circuit applies a set of trainable quantum rotation gates (\texttt{Rot} gates) to each qubit. These act like learnable weights in a neural network, letting the model adjust and transform the quantum states as needed. After that, $CRX$ gates are used to entangle the qubits in a ring-like pattern, connecting each qubit to its neighbor. This entanglement step helps the circuit capture relationships between features across the entire input.

\begin{figure}[!t]
    \centering
    \begin{subfigure}[b]{0.49\columnwidth}
        \centering
        \includegraphics[width=\linewidth]{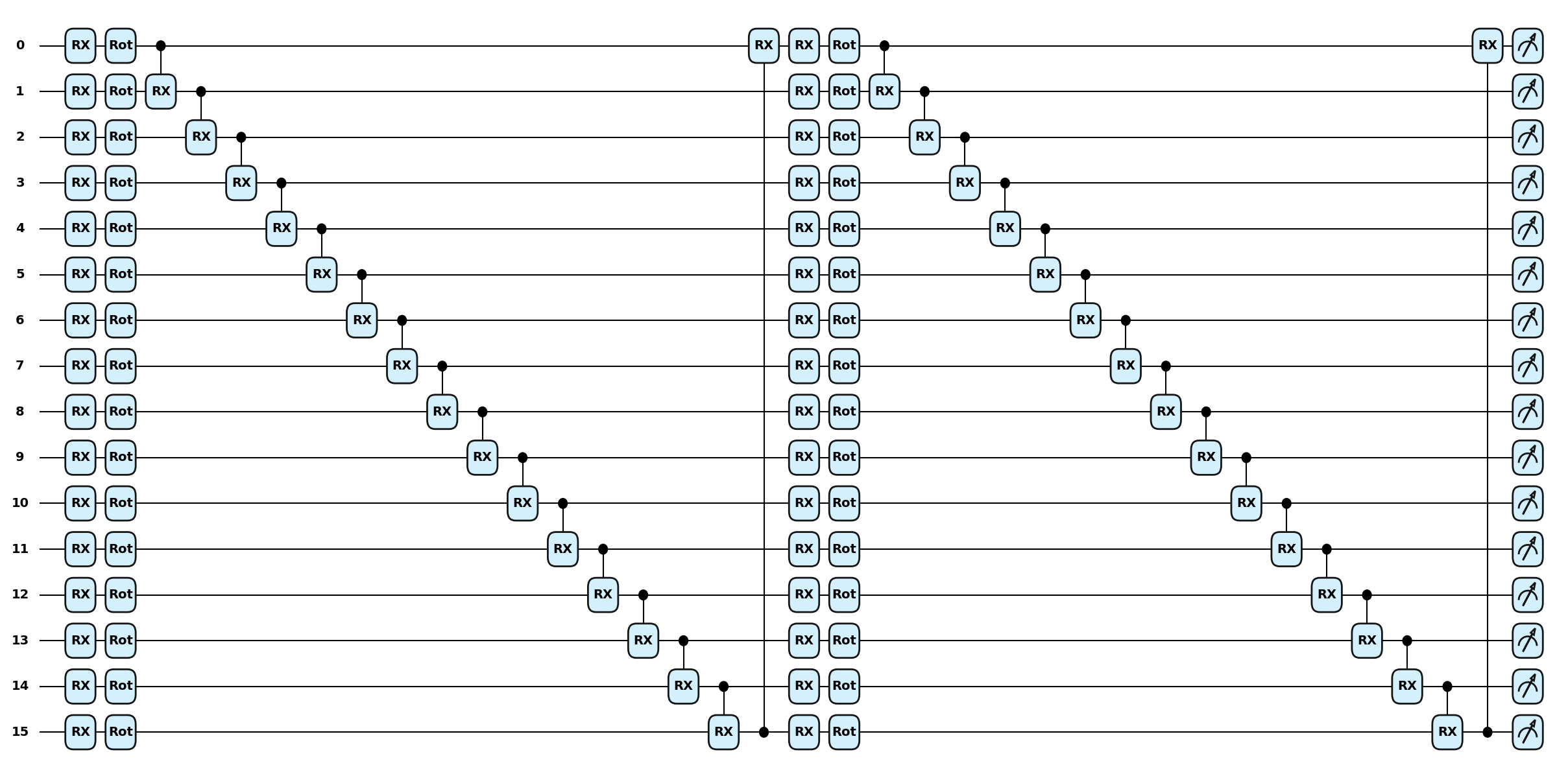}
        \caption{QMLP circuit.}
        \label{fig:qmlp}
    \end{subfigure}
    \hfill
    \begin{subfigure}[b]{0.47\columnwidth}
        \centering
        \includegraphics[width=\linewidth]{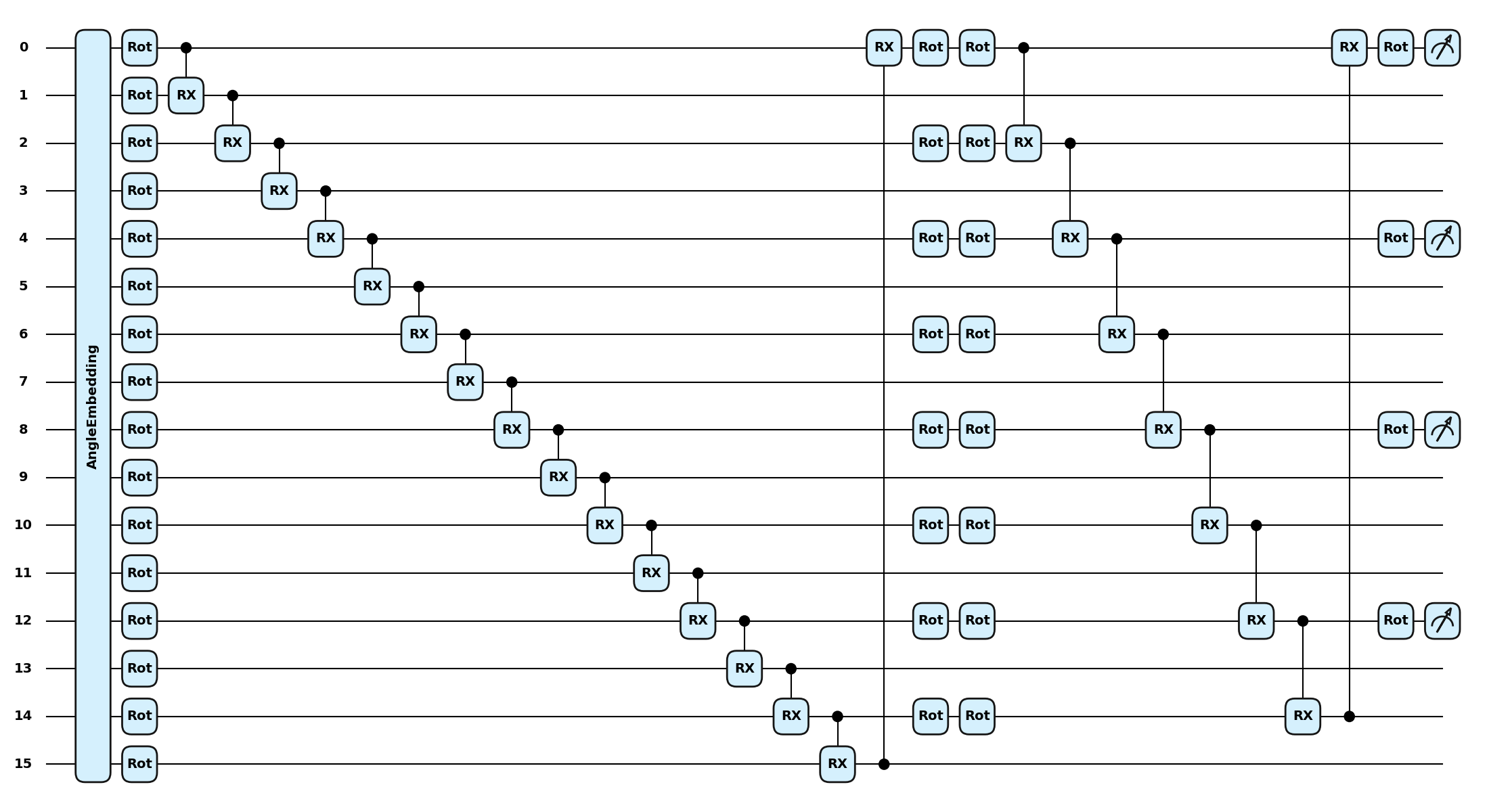}
        \caption{QCNN circuit.}
        \label{fig:qcnn}
    \end{subfigure}
    \vspace{-0.1cm}
    \caption{Quantum circuit architectures used in QMLP and QCNN models.}
    \label{fig:qmlp_qcnn}
    \vspace{-0.2cm}
\end{figure}

\begin{figure}[!t]
    \centering
    \includegraphics[width=\columnwidth]{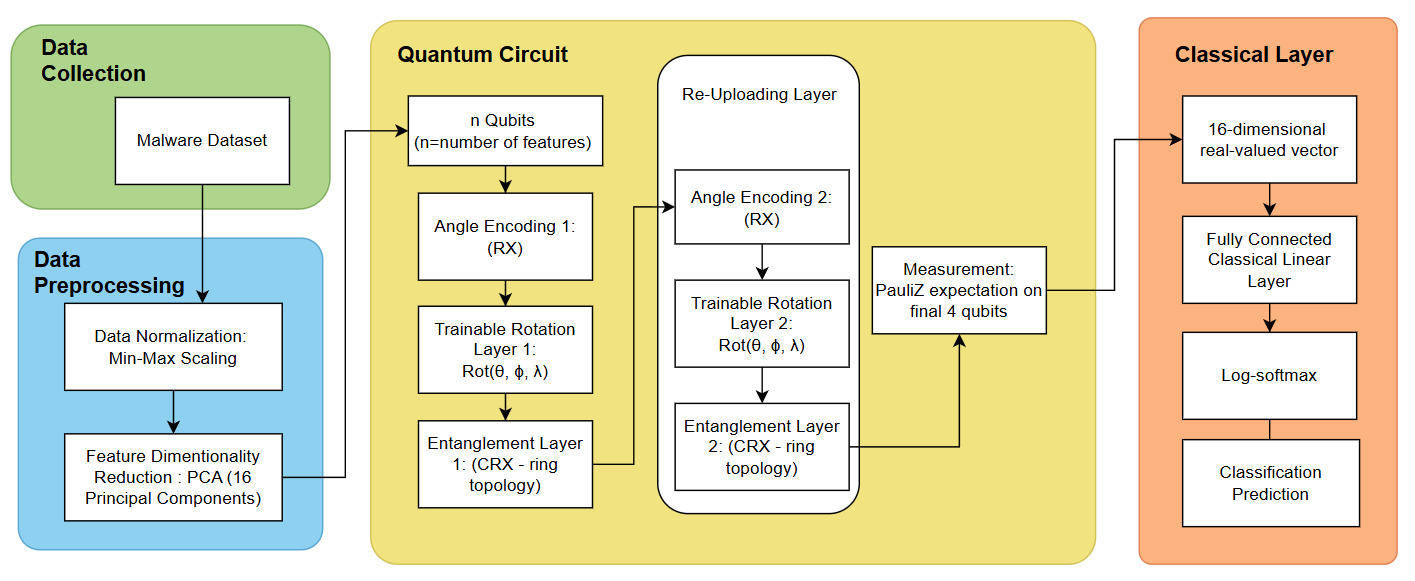}
    \vspace{-0.3cm}
    \caption{Workflow diagram of the Quantum Multilayer Perceptron (QMLP) model. Malware datasets are first normalized and reduced using Principal Component Analysis (PCA) to match the quantum input size. The resulting features are embedded into a quantum state using angle embedding. The quantum circuit then applies trainable rotation gates followed by qubit entanglement (e.g., CRX gates) and data re-uploading to enhance expressivity. After measurement using the Pauli-Z observable, the output is passed to a classical neural network layer for prediction, resulting in the final malware class labels.}
    \label{fig:qmlp-flow}
    \vspace{-0.2cm}
\end{figure}

To improve the model’s capacity to learn deeper patterns, this whole process --- $RX$ encoding, rotations, and $CRX$ entanglement, is repeated a second time with a new set of parameters. The final step in the quantum circuit is to measure the Pauli-Z expectation values from all 16 qubits, which results in a 16-dimensional feature vector that represents the learned quantum information. Figure~\ref{fig:qmlp} depicts the quantum circuit used for the QMLP model and Figure~\ref{fig:qmlp-flow} depicts the workflow of the QMLP. This quantum-generated vector is then passed into a classical linear layer, which maps it to the target class space. A log-softmax function is applied to generate the final class probabilities. To make the model compatible with batch processing in PyTorch, we use a custom \texttt{QLayer} wrapper that runs the quantum circuit on each sample in the batch and stacks the results together. 
Because of PennyLane’s integration with PyTorch through the \texttt{TorchLayer} interface, we are able to train the entire model end-to-end using Adam optimization function. This approach allows us to take advantage of quantum entanglement and nonlinearity, while still benefiting from the scalability and tools of classical deep learning.


\subsection{QCNN} 

\begin{figure}[!t]
    \centering
    \includegraphics[width=\columnwidth]{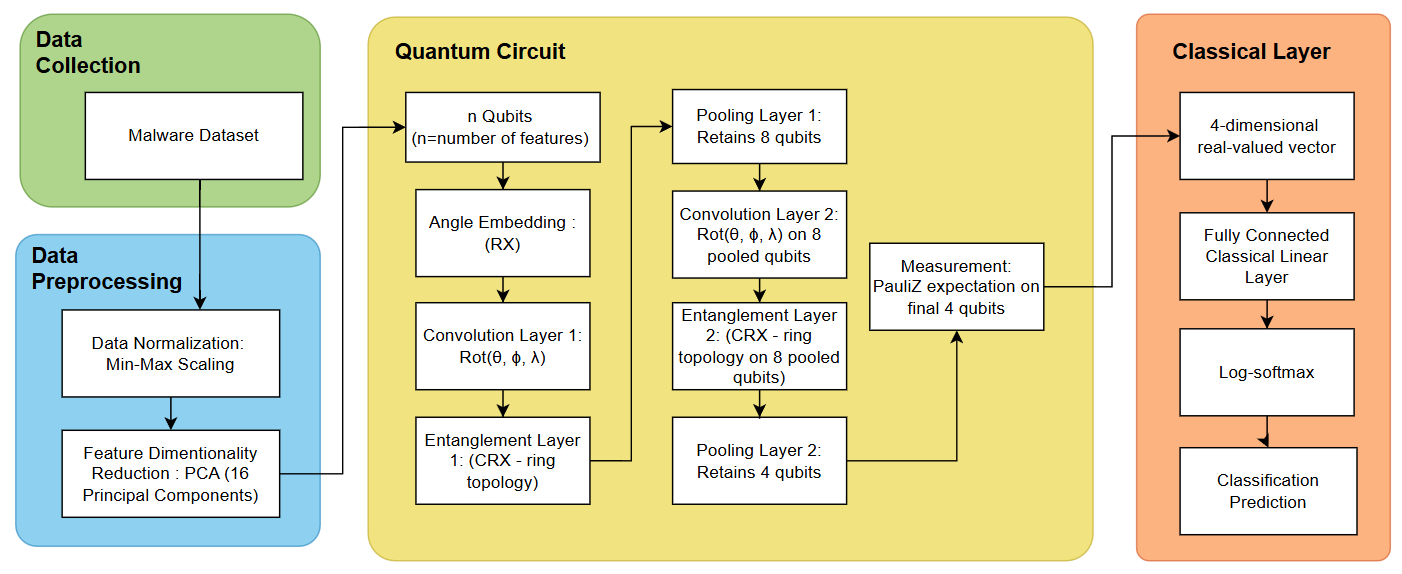}
    \vspace{-0.3cm}
    \caption{
    The QCNN workflow begins with malware datasets normalized and reduced via PCA to match the quantum input dimension. Features are embedded using angle encoding, followed by a quantum circuit with trainable rotations, CRX-based entanglement, and pooling to extract hierarchical features. This pattern is repeated, and the final qubit states are measured using the Pauli-Z observable. The outputs are then passed to a classical neural network for malware classification.
    }
    \label{fig:qcnn-flow}
    \vspace{-0.2cm}
\end{figure}

We develop a hybrid quantum-classical convolutional neural network (QCNN) that leverages quantum feature extraction capabilities with a classical machine learning classification framework \cite{arthur2022hybridquantumclassicalneuralnetwork}. The primary component of this model is a sixteen-qubit parameterized quantum circuit where each qubit represents one of the sixteen principal components derived from the malware datasets. Figure~\ref{fig:qcnn} shows the depiction of this quantum circuit and the workflow of QCNN is depicted in Figure~\ref{fig:qcnn-flow}.

To encode classical data into quantum states, quantum embedding techniques such as basis, amplitude, and angle encoding are used~\cite{Benchmarkingadversariallyrobustquantummachinelearningatscale}. This research adopts angle encoding for its simplicity in quantum layer design. It maps each input feature to the Bloch sphere using sequential single-qubit rotations, \( RX(x_i) \), \( RY(x_i) \), and \( RZ(x_i) \), providing a richer and more expressive state preparation than single-axis encoding. After feature embedding, we mimic the classical model by applying a quantum convolution layer, This layer consists of parameterized rotational gates represented as $\textit{Rot}(\theta, \phi, \lambda)$. These three parameters are learnable during training and allow each qubit’s state to be flexibly transformed. These parameters are analogous to weights in classical machine learning.  The use of different axes ensures richer quantum state manipulation, enabling the network to explore a more expressive hypothesis space.



Following the $\textit{Rot}(\theta, \phi, \lambda)$ operation, entanglement is introduced to enable interaction between the qubits. We use controlled rotation gates, specifically, the CRX gates to entangle the qubits \cite{stein2022qucnnquantumconvolutional}, with the control qubit modulates the application of an $RX$ rotation on the target qubit. The rotation angle $\alpha$ is trainable, and the entanglement follows a ring topology, connecting the last qubit back to the first to capture local correlations—analogous to convolutional layers in classical CNNs. A subsequent quantum pooling layer retains only the even-indexed qubits (0, 2, 4, ..., 14), reducing dimensionality similarly to classical pooling layers.



Each of the retained qubits then undergoes an additional learnable transformation via a parameterized single-qubit rotation gate, \textit{Rot(\( \theta, \phi, \lambda \))} and subsequently, the circuit applies another set entanglement gates to the eight retained qubits. This localized entanglement ensures that correlations among pooled features are preserved and extended, allowing for complex interactions and dependencies to emerge within the compressed quantum space. The second entanglement layer, in this way, deepens the network’s expressivity while keeping the number of active qubits. After this, another pooling layer is applied on the retained entangled qubits resulting in retaining only four qubits.

The final step in the QCNN model is to measure these final retained qubits.  This measurement is done in the computational basis to generate the final output from the quantum circuit. The expectation values of the Pauli-Z operator are extracted from these four qubits (e.g., qubits 0,4,8 and 12). These four scalar values form a fixed-length, real-valued feature vector that captures the learned quantum representation of the input data. This vector serves as the interface between the quantum and classical components of the model.

\section{Experiments}
This section details the experimental setup used to implement and evaluate the proposed Quantum Convolutional Neural Network (QCNN) and Quantum Multilayer Perceptron (QMLP). We evaluate the performance of QCNN and QMLP models using five large-scale malware datasets --- EMBER-Domain~\cite{ember}, EMBER-Class~\cite{continual-learning-malware}, AZ-Domain~\cite{androzoo,madar}, AZ-Class~\cite{malcl}, and API-Graph~\cite{api_graph_dataset} in both multiclass and binary classification settings. Our methodological approach encompasses three main phases: data preprocessing, hybrid quantum-classical model design, and evaluation.

\subsection{Dataset}
The datasets used in this study consist of static malware feature vectors with high-dimensional input spaces. For Binary Classification, the API-Graph~\cite{api_graph_dataset} dataset has 79,101 training and 58,485 testing samples with 1,159 features. The AZ-Domain~\cite{androzoo,madar} dataset has 67,656 training and 68,330 testing samples with a total number of 1,789 features. The EMBER-Domain~\cite{ember} dataset has 141,817 training samples and 75,006 testing samples, with a total number of 2,381 features. 

For multiclass classification, the API-Graph dataset for top 4 families comprised 75,485 training and 55,178 testing samples; top 14 families included 77,626 training and 56,990 testing samples; top 23 families had 78,235 training and 57,563 testing samples, all with a total of 1,159 features per sample. The AZ-Class~\cite{malcl} dataset for top 4 families 32,995 training and 4,000 testing samples; top 14 families, 125,885 training and 13,988 testing samples; top 23 families has 184,342 training and 20,482 testing samples, with all three configurations having 2,439 features per sample. The EMBER-Class~\cite{continual-learning-malware} for top 4 families has 89,281 training and 9,930 testing samples; top 14 families has 184,700 training and 20,464 testing samples; top 23 families has 220,412 training and 24,445 testing samples, with 2,381 number of features per sample. A standardized preprocessing pipeline is applied, beginning with Min-Max normalization to scale features between 0 and 1. Principal Component Analysis (PCA)~\cite{qmlp} is then used to reduce dimensionality to 16, aligning with the number of qubits in the quantum model. This reduction enables efficient quantum encoding while preserving most of the dataset's variance.



\subsection{Experimental Setup}

The hybrid models consist of classical and quantum components. The classical part is implemented using PyTorch, while the quantum circuits are built with PennyLane~\cite{PennyLane}. For the QCNN, we design a two-layer quantum circuit with entanglement, using angle embedding for feature encoding and mimicking classical convolution and pooling layers. The circuit starts with 16 qubits; after the first pooling layer, this is reduced to 8, followed by entanglement and a second pooling layer that reduces the qubits to 4. These are measured to produce the output. 
The QMLP also adopts a two-layer design with entanglement. It uses angle embedding and Rot gates to introduce trainable parameters. After the first layer, data re-uploading is applied by repeating the Rot($\theta, \phi, \lambda$) gates on all 16 qubits. All qubits are then measured to generate the output. 

In the classical postprocessing stage, the measured quantum outputs are passed through a fully connected linear layer that maps to the number of target classes, followed by a log-softmax activation. Training is done end-to-end using stochastic gradient descent, with gradients computed via PennyLane’s automatic differentiation through the \texttt{qnode}.

\begin{table*}[!t]
\centering
\scriptsize
\caption{\textbf{Summary of Results.} {Multi class classification with Quantum Multilayer Perceptron (QMLP) and Quantum Convolutional Neural Network (QCNN)}.}
\label{tab:sumMulti}
\begin{tabular}{c|l|ccc|ccc|ccc}
\toprule
\multirow{2}{*}{\textbf{Model}} & \multirow{2}{*}{\textbf{Metrics}} &  \multicolumn{3}{c|}{\textbf{API-Graph}} & \multicolumn{3}{c}{\textbf{AZ-Class}}&\multicolumn{3}{c}{\textbf{EMBER-Class}} \\ \cline{3-11}

 &  & 4 Class & 14 Class & 23 Class & 4 Class & 14 Class & 23 Class & 4 Class & 14 Class & 23 Class\\
\midrule
\multirow{6}{*}{\textbf{QMLP}} 
&Accuracy
& 95.7$\pm$0.2 & 92.5$\pm$0.1& 91.6$\pm$0.0
& 93.6$\pm$0.8 & 67.5$\pm$0.3 & 57.8$\pm$0.2 
& 88.1$\pm$0.3 & 83.9$\pm$0.2 & 79.8$\pm$0.3 \\

&Precision
& 30.1$\pm$3.5 & 6.9$\pm$0.3&4.1$\pm$0.1 
& 93.6$\pm$0.7 & 67.5$\pm$0.2 & 55.9$\pm$0.2
& 86.1$\pm$0.3 &82.9 $\pm$0.1 & 76.2$\pm$0.7 \\

&Recall
& 29.2$\pm$2 & 7.2$\pm$0.1 & 4.4$\pm$0.0
& 93.6$\pm$2 & 67.5$\pm$0.3 & 57.6$\pm$0.3
& 86.3$\pm$0.3 & 83$\pm$0.1 & 77$\pm$0.4 \\

&F1 Score
& 29.6$\pm$2.7 & 7.1$\pm$0.2 & 4.2$\pm$0.1 
& 93.6$\pm$0.7 & 67.5$\pm$0.2 & 56.7$\pm$0.2
& 86.2$\pm$0.3 & 83.0$\pm$0.1 & 76.6$\pm$0.6 \\ 

&FPR
& 15.5$\pm$2.1 & 6.6$\pm$0.2 & 4.1$\pm$0.4
& 2.1$\pm$0.2 & 2.5$\pm$0 & 1.9$\pm$0 
& 3.8$\pm$0.1 & 1.3$\pm$0 & 0.9$\pm$0 \\ 

&FNR
& 70.9$\pm$2.1 & 92.8$\pm$0.1 & 95.7$\pm$0
& 6.4$\pm$0.8 & 32.5$\pm$0.3 & 42.5$\pm$0.3 
& 13.7$\pm$0.3 & 17.0$\pm$0.1 & 23$\pm$0.4 \\ 

&ROC-AUC
& 87.6$\pm$1.9 & 85.8$\pm$2.2 & 84.7$\pm$2.4 
& 99.1$\pm$0.1 & 95.9$\pm$0.1 & 94.5$\pm$0
&97.9 $\pm$0.1 &97.5 $\pm$0 & 97.1$\pm$0.1 \\

\midrule
\multirow{6}{*}{\textbf{QCNN}}
&Accuracy
& 95.7$\pm$0.1 & 92.5$\pm$0.1 & 91.6$\pm$0.0
& 92.1$\pm$0.4 & 55.3$\pm$0.8 & 41.7$\pm$ 1.6
& 85.8$\pm$0.6 & 72.7$\pm$1.4 &60.7 $\pm$0.8 \\

&Precision
& 40.6$\pm$6 & 7.6$\pm$1.7 & 4.6$\pm$0.7 
& 92.1$\pm$0.4 & 54.2$\pm$0.9 & 36.1$\pm$0.6 
& 83.7$\pm$0.5 & 69.3$\pm$3 & 46.7$\pm$1.4 \\

&Recall
& 31.9$\pm$1.9 & 7.7$\pm$1 & 4.4$\pm$0
& 92.1$\pm$0.4 & 55.3$\pm$0.8 & 40.5$\pm$1.3
& 83.7$\pm$0.6 &67.5 $\pm$0.6 &49.7 $\pm$2.8\\

&F1 Score
& 35.5$\pm$1.8 & 7.6$\pm$1.3 & 4.49$\pm$0.3
& 92.1$\pm$0.4 & 54.8$\pm$0.9 & 38.19$\pm$0.9 
& 83.7$\pm$0.5 & 68.4$\pm$1.5 & 48.11$\pm$1.84 \\ 

&FPR
& 19.6$\pm$3.1 & 6.3$\pm$1 & 4.2$\pm$0.2
& 2.6$\pm$0.1 & 3.4$\pm$0 & 2.7$\pm$0 
& 4.5$\pm$0.2 & 2.1$\pm$0.1 & 1.83$\pm$0 \\ 

&FNR
& 68.1$\pm$1.9 & 92.3$\pm$1 & 95.6$\pm$0
& 7.9$\pm$0.4 & 44.7$\pm$0.8 & 59.5$\pm$1.3 
&16.3 $\pm$0.6 & 32.5$\pm$0.6 & 50.3$\pm$2.8 \\ 

&ROC-AUC& 90.5$\pm$1.8 & 82$\pm$1.8 & 78.9$\pm$5.00 
& 98.6$\pm$0.1 & 92.9$\pm$0.1 & 89$\pm$0.3 
& 97.1$\pm$0.1 & 94$\pm$0.2 & 92.5$\pm$0.3 \\
\bottomrule
\end{tabular}
\vspace{-0.3cm}
\end{table*}

\begin{table}[!t]
\centering
\scriptsize
\caption{\textbf{Summary of Results.} {Binary Classification with QMLP and QCNN}.}
\label{tab:sumBinary}
\begin{tabular}{c|l|c|c|c}
\toprule
\textbf{Model} & \textbf{Metrics} &  \multicolumn{1}{c|}{\textbf{API-Graph}} & \multicolumn{1}{c|}{\textbf{AZ-Domain}}&\multicolumn{1}{c}{\textbf{EMBER-Domain}} \\
\midrule
\multirow{6}{*}{\textbf{QMLP}} &Accuracy&96.3$\pm$0.2&91.24$\pm$0.3&77.4$\pm$2.5\\
&Precision& 91.0$\pm$0.5 & 76.64$\pm$1.2 & 77.3$\pm$2.5 \\
&Recall& 86.5$\pm$1.4 & 69$\pm$0.7 & 77.2$\pm$2.5 \\
&F1 Score& 88.7$\pm$0.6 & 72.6$\pm$0.5 & 77.3$\pm$2.5 \\
&FPR& 13.5$\pm$1.4 & 31$\pm$0.7 & 22.8$\pm$2.5 \\
&FNR& 13.5$\pm$1.4 & 31$\pm$0.7 & 23.0$\pm$2.5 \\ 
&ROC-AUC& 97.0$\pm$0.2 &86.2$\pm$2.2 & 83.0$\pm$2.3 \\
\midrule
\multirow{6}{*}{\textbf{QCNN}}
&Accuracy&96.0$\pm$0.1 & 91.3$\pm$0.2 & 77.5$\pm$0.2 \\
&Precision& 91.0$\pm$0.3 & 77$\pm$0.9 & 77.4$\pm$0.1 \\
&Recall& 83.8$\pm$0.3 & 69$\pm$0.7 & 77.3$\pm$0.3 \\
&F1 Score& 87.3$\pm$0.2 & 72.8$\pm$0.7 & 77.4$\pm$0.2 \\
&FPR& 16.2$\pm$0.3 & 31$\pm$0.7 & 22.7$\pm$0.3 \\
&FNR& 16.2$\pm$0.3 & 31$\pm$0.7 & 22.7$\pm$0.3 \\ 
&ROC-AUC& 97.1$\pm$0.1 & 86$\pm$0.8 & 84.1$\pm$0.4 \\
\bottomrule
\end{tabular}
\vspace{-0.5cm}
\end{table}

\subsection{Training and Evaluation} 

For binary classification with AZ-Domain dataset, we use training data from 2008 to 2010, while evaluation is performed on samples from 2011 to 2016. The API-Graph dataset includes training data from 2012–2013 and evaluated in data from 2014. For EMBER-Domain, models are trained on samples from the first three months of 2018 and evaluated on data from the rest of months of 2018.


For multiclass classification, we use selected subsets of malware families from the API-Graph, EMBER-Class, and AZ-Class datasets. To ensure consistency, we fix the number of classes to 4, 14, and 23 families respectively across the three datasets. Each experiment, binary and multiclass, is repeated three times per model. Both quantum models are simulated using PennyLane's~\cite{PennyLane} \texttt{default.qubit} backend, which provides ideal (noise-free) quantum outputs. Each model is trained for 20 epochs with a batch size of 64 using 16-qubit circuits. The final results are reported as the average across these independent runs to capture model consistency and variability.



\section{Results}
This section presents the malware classification results of QCNN and QMLP. Performance is evaluated using accuracy, macro-averaged precision, recall, F1-score, false positive rate (FPR), false negative rate (FNR), and ROC-AUC.



%
%

For binary classification, Table \ref{tab:sumBinary} summarizes the performance metrics for both QMLP and QCNN models. In API-Graph dataset, QMLP had an accuracy of 96.3\% while the QCNN had 95.9\%. Both models achieved an identical precision of 91\%. Recall was somewhat higher in QMLP with 86.5\% while in QCNN we had a recall of 83.8\%. 
For AZ-Domain, we observe that the accuracy was somewhat consistent on both QML models, being 91.71\% and 91.8\% for QMLP and QCNN respectively. Other metrics remained quite close; for precision, QCNN was slightly better with 80.2\% while QMLP 79.9\%, and for F1 Score, QCNN had 75.9\% compared to the 75.8\%. Notably, both models exhibited their highest False Positive Rate (FPR) and False Negative Rate (FNR) on the AZ-Domain dataset, reaching approximately 27.8\% and 28\%, respectively.
On the EMBER-Domain, both QMLP and QCNN models exhibited very similar average performance for Accuracy, Precision, recall and F1 score varying from 77.2\% to 77.4\%. The most noticeable variation, however, was in their standard deviations: QMLP consistently showed a standard deviation of $\pm$2.5 across all metrics, compared to QCNN's, which ranged from $\pm$0.1 to $\pm$0.4.

%
%

Table \ref{tab:sumMulti} summarizes the performance metrics for the QMLP and QCNN models in multiclass classification. For the API-Graph dataset, both models exhibited identical accuracies of 95.7\%, 92.5\% and 91.6\% for 4,14 and 23 class configurations, respectively. QCNN demonstrated better precision and F1 score across the three configurations of API-Graph dataset. Despite the high accuracy values, both models showed remarkably low precision, recall, and F1 scores. These metrics consistently dropped across configurations with values reaching as low as 4.1\% for Precision, 4.4\% for Recall, and 4.2\% for F1-Score in the 23-class configurations.

On the AZ-Class dataset, the QMLP model exhibited a significant reduction in performance as the number of classes is increased. Its Accuracy, Precision, Recall and F1 Score, which uniformly started at 93.6\% for the 4-class configuration, approximately ended in the range of 55.9\% to 57.8\% for the 23-class configuration. Similarly, the QCNN model also showed a substantial performance decline, its Accuracy, Precision, Recall, F1-Score all began at 92.1\% for the 4-class configuration and ended in a range from 36.1\% to 41.7 for the 23-class configuration.

For EMBER-Class dataset, as the number of classes increase, the QMLP experience Accuracy reduction of approximately 10\%, from 88.1\% (4-class) to 79.8\% (23-class). Similar downward trends were observed for Precision, Recall and F1 score. The False negative rate (FNR), in contrast, increased by almost 10\% with class expansion, rising from 13.7\% in the 4-class configuration to 23\% in the 23-class configuration. On the other hand, the QCNN experienced a more substantial reduction in accuracy, dropping by approximately 25\% from 85.8\% (4-class) to 60.7\% (23-class). In addition, its Precision, Recall and F1 Score dropped by almost half, starting approximately on 83.7\% for all the three configurations and concluding on 46.7\% (Precision), 49.7\% (Recall) and 48.1\% (F1-Score) respectively in the 23-class configuration.



\section{Discussion}
\paragraphX{\textbf{QMLP vs. QCNN Performance.}} The results indicate that the QMLP generally outperforms the QCNN in multiclass classification, particularly as the number of classes increases. QMLP achieves higher accuracy, stronger F1 scores, and lower false negative rates on the AZ-Class and EMBER-Class datasets. In binary classification tasks, both models perform well, achieving high ROC-AUC scores across all datasets. On the API-Graph dataset, QMLP maintains a slight edge in recall and F1 score—likely due to its architecture, which measures all qubits and incorporates data re-uploading to enhance representational capacity.

\paragraphX{\textbf{Efficiency.}} QMLP is more expensive to train due to the overhead of repeated qubit measurements and re-uploaded parameters. In contrast, QCNN benefits from faster training through a pooling mechanism that reduces the number of active qubits at each layer. While this improves efficiency, it also limits circuit expressivity, as only half of the qubits are retained per layer. This reduction may hinder the model's ability to capture complex decision boundaries, particularly in multiclass settings.

\paragraphX{\textbf{QCNN Scalability Challenges.}} QCNN shows stable performance in binary tasks. However, its classification accuracy degrades significantly in more complex multiclass scenarios dropping by 36.8\% and 50.4\% on AZ-Class, and by 13.1\% and 25.1\% on EMBER-Class for 14 and 23 classes, respectively. These declines highlight QCNN’s limitations in handling high class diversity, underscoring the need for more expressive quantum architectures in such settings.

\paragraphX{\textbf{Dimensionality Constraint.}} A design constraint in this study involves reducing input dimensionality to 16 principal components using PCA. This was necessary to apply Angle Embedding as a feature encoder. While simulating models with more than 16 qubits is technically feasible, this constraint is imposed to maintain uniformity across experiments and to ensure practical runtimes during the quantum simulations. An alternative direction for future research could be the use of Amplitude Embedding, which is more flexible by allowing the encoding of $2^n$ features using $n$ qubits.

\paragraphX{\textbf{Simulation Assumptions.}} Lastly, it is important to note that all experiments are conducted under ideal, noise-free conditions using PennyLane’s \texttt{default.qubit} simulator. This backend does not model real-world quantum noise, such as decoherence or gate errors. While useful for controlled evaluations, such idealized simulations may not reflect performance on actual quantum hardware. A promising future direction involves incorporating noise-aware simulators—such as IBM’s \texttt{qiskit\_aer}—to more realistically assess robustness and feasibility in near-term quantum devices.

\section{Related Work}

\subsection{Quantum Machine Learning for Computer Vision}

Prior research into Quantum Machine Learning (QML), more specifically quantum neural networks (QNNs), have focused on improving inference accuracy and performance \cite{hur2021quantum}, noise robustness \cite{mittal2022nrqnn}, and nonlinearity \cite{moreira2023realization}. Early QNN models such as QuantumFlow \cite{jiang2021} introduce basic neuron-like behavior using fixed gates and binary weights. Due to the large circuit depth, QuantumFlow perform poorly under noise; its inference accuracy dropped to around 10\% in noisy environments, reducing it to random guessing. QuantumNAS\cite{wang2022}, a more sophisticated QNN applied neural architecture search to automatically create variational quantum circuits from a large design space to improve performance. However, it lacks error-awareness during training and relies on entanglement-based encoding, which amplifies noise propagation, thereby limiting its scalability and robustness on real quantum hardware.

To address these challenges, Chu et al.~\cite{qmlp} proposes the Quantum Multi-Layer Perceptron (QMLP) tailored for Noisy Intermediate-Scale Quantum (NISQ) devices. QMLP employs error-tolerant input encoding by isolating features on separate qubits using $RX$ gates, reducing the impact of single-gate errors. It introduces re-uploading units to add tunable nonlinearity through repeated data injection, mimicking classical activation functions. QMLP achieves up to 10\% accuracy gain on the MNIST dataset and exhibits greater robustness to quantum errors such as bit-flips and phase-flips.

Despite recent advancements, QNNs still suffer from several limitations most notably, {\em barren plateaus}. A barren plateau arises when the quantum model’s training gradients vanish exponentially due to too much expressiveness, entanglement, or nonlocality — as formalized in~\cite{ragone2024lie}. Quantum Convolutional Neural Networks (QCNNs) have demonstrated the potential to avoid Barren Plateaus which supports the scalability of quantum models. This claim is supported by Pesah et al \cite{pesah2021absence}, who proved that QCNNs avoid barren plateaus by showing that their cost function gradient vanish only polynomially with system size even under random initialization. 


\subsection{Quantum Machine Learning for Malware Analysis}

While image classification has primarily served as a benchmark for QNN development as seen in \cite{qmlp, qnn2022evaluation}, malware analysis presents unique challenges that go beyond image complexity. For example, Mercaldo et al. \cite{mercaldo2022towards} demonstrated that Android malware images converted from $.dex$ bytecode retained structured patterns that could be effectively leveraged by both classical and quantum classifiers for family-level malware detection. These images, though visually noisy, preserved semantic traits derived from the underlying bytecode, enabling models to distinguish between benign and malicious applications. Furthermore, the authors used Grad-CAM to show that the models focused on consistent, meaningful image regions and thus validating that these visual representations encoded learnable malware behavior rather than random artifacts. In other words, effective malware detection requires attention to the semantic layout of executable formats, and is sensitive to adversarial obfuscation and section manipulation. For example, the $.rsrc$ (resource) section is often exploited by malware to evade detection; scripts can inject payloads directly into this section, which are later extracted and executed when the binary runs \cite{quertier2023malware}. Therefore, techniques optimized for image classification may not directly translate to robust malware classification without domain-specific adjustments.

Quertier et al. \cite{quertier2023malware} applies QCNNs to the task of malware classification using Portable Executable (PE) files. They propose a distributed QCNN framework where each QCNN is trained on images generated from specific PE sections (e.g., $.text$, $.data$, $.rsrc$). Using a grayscale transformation and section-specific training, their approach overcame the qubit limitations of NISQ\cite{Preskill2018} devices by decomposing the binary file into smaller, more manageable sub-images. To integrate the outputs, a scoring function (e.g., XGBoost) aggregated the section-wise predictions, resulting in a roughly 20\% improvement in both accuracy and F1-score over single QCNN models trained on whole-file images. Their work demonstrates the potential of hybrid quantum-classical architectures in security-related applications, especially where structured data and limited qubit resources present major obstacles.

\section{conclusion}

In this study, we explore quantum machine learning models for malware classification. QMLP outperforms QCNN across datasets, benefiting from full-qubit measurement and data re-uploading for enhanced pattern learning. However, limitations remain, including the high computational cost of quantum simulation, reliance on ideal noise-free conditions via PennyLane’s \texttt{default.qubit}, and potential information loss from dimensionality reduction. We will explore noise-aware quantum simulators and improved techniques to preserve information during dimensionality reduction in the future work.


\bibliographystyle{IEEEtran}
\bibliography{ref}

\begin{thebibliography}{10}
\providecommand{\url}[1]{#1}
\csname url@samestyle\endcsname
\providecommand{\newblock}{\relax}
\providecommand{\bibinfo}[2]{#2}
\providecommand{\BIBentrySTDinterwordspacing}{\spaceskip=0pt\relax}
\providecommand{\BIBentryALTinterwordstretchfactor}{4}
\providecommand{\BIBentryALTinterwordspacing}{\spaceskip=\fontdimen2\font plus
\BIBentryALTinterwordstretchfactor\fontdimen3\font minus \fontdimen4\font\relax}
\providecommand{\BIBforeignlanguage}[2]{{%
\expandafter\ifx\csname l@#1\endcsname\relax
\typeout{** WARNING: IEEEtran.bst: No hyphenation pattern has been}%
\typeout{** loaded for the language `#1'. Using the pattern for}%
\typeout{** the default language instead.}%
\else
\language=\csname l@#1\endcsname
\fi
#2}}
\providecommand{\BIBdecl}{\relax}
\BIBdecl

\bibitem{nguyen2025diffusion}
H.-Q. Nguyen, X.~B. Nguyen, S.~Y.-C. Chen, H.~Churchill, N.~Borys, S.~U. Khan, and K.~Luu, ``Diffusion-inspired quantum noise mitigation in parameterized quantum circuits,'' \emph{Quantum Machine Intelligence}, 2025.

\bibitem{cerezo2022challenges}
M.~Cerezo, G.~Verdon, H.-Y. Huang, L.~Cincio, and P.~J. Coles, ``Challenges and opportunities in quantum machine learning,'' \emph{Nature Computational Science}, 2022.

\bibitem{biamonte2017quantum}
J.~Biamonte, P.~Wittek, N.~Pancotti, P.~Rebentrost, N.~Wiebe, and S.~Lloyd, ``Quantum machine learning,'' \emph{Nature}, 2017.

\bibitem{schuld2014quest}
M.~Schuld, I.~Sinayskiy, and F.~Petruccione, ``The quest for a quantum neural network,'' \emph{Quantum Information Processing}, 2014.

\bibitem{abbas2021power}
A.~Abbas, D.~Sutter, C.~Zoufal, A.~Lucchi, A.~Figalli, and S.~Woerner, ``The power of quantum neural networks,'' \emph{Nature Computational Science}, 2021.

\bibitem{Preskill2018}
J.~Preskill, ``Quantum computing in the nisq era and beyond,'' \emph{Quantum}, 2018.

\bibitem{hur2021quantum}
H.-S. Hur, J.~Kim, and J.~Kim, ``Quantum convolutional neural network for classical data classification,'' \emph{arXiv preprint arXiv:2108.00661}, 2021.

\bibitem{wang2022}
H.~Wang, Y.~Ding, J.~Gu, Y.~Lin, D.~Z. Pan, F.~T. Chong, and S.~Han, ``{QuantumNAS}: Noise-adaptive search for robust quantum circuits,'' in \emph{2022 IEEE International Symposium on High-Performance Computer Architecture (HPCA)}, 2022.

\bibitem{wang2021quantumnat}
H.~Wang, J.~Gu, Y.~Ding, Z.~Li, F.~T. Chong, D.~Z. Pan, and S.~Han, ``{QuantumNAT}: Quantum noise-aware training with noise injection, quantization and normalization,'' \emph{arXiv preprint arXiv:2110.11331}, 2021, accessed 2025-06-16.

\bibitem{qnn2022evaluation}
M.~S. Rahman, N.~Ishita, S.~Shihab, and K.~Vatanparvar, ``An evaluation of hardware-efficient quantum neural networks,'' \emph{Electronics}, 2022.

\bibitem{PennyLane}
V.~Bergholm, J.~Izaac, M.~Schuld, C.~Gogolin, S.~Ahmed, V.~Ajith, M.~S. Alam, G.~Alonso-Linaje, B.~AkashNarayanan, A.~Asadi \emph{et~al.}, ``{PennyLane}: Automatic differentiation of hybrid quantum-classical computations,'' \emph{arXiv preprint arXiv:1811.04968}, 2018.

\bibitem{tensorflow-quantum}
M.~Broughton, G.~Verdon, T.~McCourt, A.~J. Martinez, J.~H. Yoo, S.~V. Isakov, P.~Massey, R.~Halavati, M.~Y. Niu, A.~Zlokapa \emph{et~al.}, ``{TensorFlow Quantum}: A software framework for quantum machine learning,'' \emph{arXiv preprint arXiv:2003.02989}, 2020.

\bibitem{qmlp}
C.~Chu, N.-H. Chia, L.~Jiang, and F.~Chen, ``{QMLP}: An error-tolerant nonlinear quantum mlp architecture using parameterized two-qubit gates,'' in \emph{Proceedings of the ACM/IEEE International Symposium on Low Power Electronics and Design}, 2022.

\bibitem{sajjan2022quantum}
M.~Sajjan, J.~Li, R.~Selvarajan, S.~H. Sureshbabu, S.~S. Kale, R.~Gupta, V.~Singh, and S.~Kais, ``Quantum machine learning for chemistry and physics,'' \emph{Chemical Society Reviews}, 2022.

\bibitem{choudhary2019accelerated}
K.~Choudhary, M.~Bercx, J.~Jiang, R.~Pachter, D.~Lamoen, and F.~Tavazza, ``Accelerated discovery of efficient solar cell materials using quantum and machine-learning methods,'' \emph{Chemistry of materials}, 2019.

\bibitem{qcnn}
I.~Cong, S.~Choi, and M.~D. Lukin, ``Quantum convolutional neural networks,'' \emph{Nature Physics}, 2019.

\bibitem{ember}
H.~S. Anderson and P.~Roth, ``{EMBER}: {An} open dataset for training static {PE} malware machine learning models,'' \emph{arXiv:1804.04637}, 2018.

\bibitem{api_graph_dataset}
X.~Zhang, Y.~Zhang, M.~Zhong, D.~Ding, Y.~Cao, Y.~Zhang, M.~Zhang, and M.~Yang, ``Enhancing state-of-the-art classifiers with api semantics to detect evolved android malware,'' in \emph{ACM Conference on Computer and Communications Security (CCS)}, 2020.

\bibitem{arp2014drebin}
D.~Arp, M.~Spreitzenbarth, M.~Hubner, H.~Gascon, K.~Rieck, and C.~Siemens, ``{Drebin}: Effective and explainable detection of android malware in your pocket,'' in \emph{Network and Distributed System Security Symposium (NDSS)}, 2014.

\bibitem{continual-learning-malware}
M.~S. Rahman, S.~E. Coull, and M.~Wright, ``On the limitations of continual learning for malware classification,'' in \emph{First Conference on Lifelong Learning Agents (CoLLAs)}, 2022.

\bibitem{malcl}
J.~Park, A.~Ji, M.~Park, M.~S. Rahman, and S.~E. Oh, ``{MalCL}: Leveraging gan-based generative replay to combat catastrophic forgetting in malware classification,'' in \emph{AAAI Conference on Artificial Intelligence (AAAI)}, 2025.

\bibitem{madar}
M.~S. Rahman, S.~Coull, Q.~Yu, and M.~Wright, ``{MADAR}: Efficient continual learning for malware analysis with diversity-aware replay,'' \emph{arXiv preprint arXiv:2502.05760}, 2025.

\bibitem{arthur2022hybridquantumclassicalneuralnetwork}
D.~Arthur and P.~Date, ``A hybrid quantum-classical neural network architecture for binary classification,'' 2022.

\bibitem{Benchmarkingadversariallyrobustquantummachinelearningatscale}
M.~T. West, S.~M. Erfani, C.~Leckie, M.~Sevior, L.~C.~L. Hollenberg, and M.~Usman, ``Benchmarking adversarially robust quantum machine learning at scale,'' \emph{Phys. Rev. Res.}, 2023.

\bibitem{stein2022qucnnquantumconvolutional}
S.~A. Stein, Y.~Mao, J.~Ang, and A.~Li, ``Qucnn : A quantum convolutional neural network with entanglement based backpropagation,'' 2022.

\bibitem{androzoo}
K.~Allix, T.~F. Bissyand{\'e}, J.~Klein, and Y.~Le~Traon, ``{{AndroZoo}: Collecting Millions of Android Apps for the Research Community},'' in \emph{International Conference on Mining Software Repositories (MSR)}, 2016.

\bibitem{mittal2022nrqnn}
Y.~Mittal, S.~Choudhary, S.~Nandi, S.~Dasgupta, and V.~Singh, ``Nr-qnn: A noise-resilient quantum neural network architecture,'' \emph{arXiv preprint arXiv:2210.06778}, 2022.

\bibitem{moreira2023realization}
R.~Moreira, S.~Rosenblum, J.~M. Martinis, M.~Mohseni, H.~Neven, V.~Smelyanskiy, J.~R. McClean, M.~Broughton, S.~Boixo, M.~P. Harrigan \emph{et~al.}, ``Realization of a quantum neural network using repeat-until-success circuits in a superconducting quantum processor,'' \emph{npj Quantum Information}, 2023.

\bibitem{jiang2021}
W.~Jiang, J.~Xiong, and Y.~Shi, ``A co-design framework of neural networks and quantum circuits towards quantum advantage,'' \emph{Nature Communications}, no.~1, p. 579, 2021.

\bibitem{ragone2024lie}
M.~Ragone, B.~N. Bakalov, F.~Sauvage, A.~F. Kemper, C.~Ortiz~Marrero, M.~Larocca, and M.~Cerezo, ``A lie algebraic theory of barren plateaus for deep parameterized quantum circuits,'' \emph{Nature Communications}, 2024.

\bibitem{pesah2021absence}
A.~Pesah, M.~Cerezo, S.~Wang, T.~Volkoff, A.~T. Sornborger, and P.~J. Coles, ``Absence of barren plateaus in quantum convolutional neural networks,'' \emph{Physical Review X}, vol.~11, no.~4, p. 041011, 2021.

\bibitem{mercaldo2022towards}
F.~Mercaldo, G.~Ciaramella, G.~Iadarola, M.~Storto, F.~Martinelli, and A.~Santone, ``Towards explainable quantum machine learning for mobile malware detection and classification,'' \emph{Applied Sciences}, 2022.

\bibitem{quertier2023malware}
T.~Quertier and G.~Barrué, ``Towards an in-depth detection of malware using distributed qcnn,'' \emph{arXiv preprint arXiv:2312.12161}, 2023.

\end{thebibliography}

\end{document}